\pdfoutput=1

\documentclass[11pt]{article}

\usepackage{ACL2023}

\usepackage{times}
\usepackage{latexsym}

\usepackage[T1]{fontenc}

\usepackage[utf8]{inputenc}

\usepackage{microtype}

\usepackage{inconsolata}

\usepackage{amsmath}
\usepackage{graphicx}
\usepackage{enumitem}

\title{Training-free Neural Architecture Search for RNNs and Transformers}

\author{Aaron Serianni, \\
  Princeton University \\
  \texttt{serianni@princeton.edu} \\\And
  Jugal Kalita \\
  University of Colorado Colorado Springs \\
  \texttt{jkalita@uccs.edu} \\}

\begin{document}
\maketitle
\begin{abstract}
    Neural architecture search (NAS) has allowed for the automatic creation of new and effective neural network architectures, offering an alternative to the laborious process of manually designing complex architectures. However, traditional NAS algorithms are slow and require immense amounts of computing power. Recent research has investigated training-free NAS metrics for image classification architectures, drastically speeding up search algorithms. In this paper, we investigate training-free NAS metrics for recurrent neural network (RNN) and BERT-based transformer architectures, targeted towards language modeling tasks. First, we develop a new training-free metric, named hidden covariance, that predicts the trained performance of an RNN architecture and significantly outperforms existing training-free metrics. We experimentally evaluate the effectiveness of the hidden covariance metric on the NAS-Bench-NLP benchmark. Second, we find that the current search space paradigm for transformer architectures is not optimized for training-free neural architecture search. Instead, a simple qualitative analysis can effectively shrink the search space to the best performing architectures. This conclusion is based on our investigation of existing training-free metrics and new metrics developed from recent transformer pruning literature, evaluated on our own benchmark of trained BERT architectures. Ultimately, our analysis shows that the architecture search space and the training-free metric must be developed together in order to achieve effective results. Our source code is available at \url{https://github.com/aaronserianni/training-free-nas}.
\end{abstract}

\section{Introduction}

Recurrent neural networks (RNNs) and BERT-based transformer models with self-attention have been extraordinarily successful in achieving state-of-the-art results on a wide variety of language modeling-based natural language processing (NLP) tasks, including question answering, sentence classification, tagging, and natural language inference~\cite{brown_language_2020, palangi_deep_2016, raffel_exploring_2020, sundermeyer_lstm_2012, yu_review_2019}. However, the manual development of new neural network architectures has become increasingly difficult as models are getting larger and more complicated. Neural architecture search (NAS) algorithms aim to procedurally design and evaluate new, efficient, and effective architectures within a predesignated search space~\cite{zoph_neural_2017}. NAS algorithms have been extensively used for developing new convolutional neural network (CNN) architectures for image classification, with many surpassing manually-designed architectures and achieving state-of-the-art results on many classification benchmarks~\cite{tan_efficientnet_2019, real_regularized_2019}. Some research has been conducted on NAS for RNNs and transformers~\cite{so_evolved_2019, so_searching_2021, jing_nasabn_2020}, particularly with BERT-based architectures~\cite{yin_autotinybert_2021, xu_nas-bert_2021, gao_autobert-zero_2022, tuli_flexibert_2022, chitty-venkata_neural_2022}, but NAS is not widely used for designing these architectures.
    
    While NAS algorithms and methods have been successful in developing novel and effective architectures, there are two main problems that current algorithms face. The search space for various architectures is immense, and the amount of time and computational power to run NAS algorithms is prohibitively expensive~\cite{mehta_nas-bench-suite_2022}. Because traditional NAS algorithms require the evaluation of candidate architectures in order to gauge performance, candidate architectures need to be trained fully, each taking days or weeks to complete. Thus, past attempts at NAS have been critiqued for being computationally resource-intensive, consuming immense amounts of electricity, and producing large amounts of carbon emissions~\cite{strubell_energy_2019}. These problems are especially true for transformers and RNNs, as they have more parameters and take longer to train when compared to other architectures~\cite{so_evolved_2019, zhou_training-free_2022}.
    
    Recently, there has been research into training-free NAS metrics and algorithms, which offer significant performance increases over traditional NAS algorithms~\cite{abdelfattah_zero-cost_2020, mellor_neural_2021, zhou_training-free_2022}. These metrics aim to partially predict an architecture's trained accuracy from its initial untrained state, given a subset of inputs. However, prior research has focused on developing training-free NAS metrics for CNNs and Vision Transformers with image classification tasks. In this work, we apply existing training-free metrics and create our own metrics for RNNs and BERT-based transformers with language modeling tasks. Our main contributions are:

    \begin{itemize}[leftmargin=*]
        \setlength{\itemsep}{-0.2em}
        \item We develop a new training-free metric for RNN architectures, called ``hidden covariance,'' which significantly outperforms existing metrics on NAS-Bench-NLP.
        \item We develop a NAS benchmark for BERT-based architectures utilizing the FlexiBERT search space and ELECTRA pretraining scheme.
        \item We evaluate existing training-free metrics on our NAS BERT benchmark, and propose a series of new metrics adapted from attention head pruning.
        \item Finally, we discuss current limitations with training-free NAS for transformers due to the structure of transformer search spaces, and propose an alternative paradigm for speeding up NAS algorithms based on scaling laws of transformer hyperparameters.
    \end{itemize}

\section{Related Work}
    Since the development and adoption of neural architecture search, there has been research into identifying well-performing architectures without the costly task of training candidate architectures.
    
\subsection{NAS Performance Predictors}
    Prior attempts at predicting a network architecture's accuracy focused on training a separate performance predictor. \citet{deng_peephole_2017} and \citet{istrate_tapas_2019} developed methods called Peephole and Tapas, respectively, to embed the layers in an untrained CNN architecture into vector representations of fixed dimension. Then, both methods trained LSTM networks on these vector representations to predict the trained architecture's accuracy. Both methods achieved strong linear correlations between the LSTMs' predicted accuracy and the actual trained accuracy of the CNN architectures. In addition, the LSTM predictors can quickly evaluate many CNN architectures. The main limitation of these methods is that the LSTM predictors require large amounts of trained CNN architectures to accurately train the predictors, thus not achieving the goal of training-free NAS. 

\subsection{Training-free Neural Architecture Search}
    \citet{mellor_neural_2021} presented a method for scoring a network architecture without any training and prior knowledge of trained network architectures. They focused on CNN architectures in the sample space of various NAS benchmarks, predicting the accuracy of the architectures on the CIFAR-10, CIFAR-100, and ImageNet image classification benchmarks. While \citeauthor{mellor_neural_2021}'s proposed method showed a correlation between their score and actual trained accuracy, it decreased with more complex datasets like ImageNet and architectures with high accuracy. \citeauthor{mellor_neural_2021} found that the images chosen for the mini-batch and initialization weights of the model have negligible impact on their score. Their method can predict accuracies of architectures in seconds, and is easily combined with traditional NAS algorithms.

    \citet{abdelfattah_zero-cost_2020} introduced a series of additional training-free metrics for CNNs with image classification tasks, based in network pruning literature, aiming to improve performance. They also tested  their metrics on other search spaces with different tasks, including NAS-Bench-NLP with RNNs and NAS-Bench-ASR, but found significantly reduced performance in these search spaces.

\section{Training-free NAS Metrics}
    A series of training-free NAS metrics have been proposed in recent literature. These metrics look at specific aspects of an architecture, such as parameter gradients, activation correlations, and weight matrix rank. Most metrics can be generalized to any type of neural network, but have only been tested on CNN architectures. For transformer architectures, we also adapt various attention parameter pruning metrics as training-free metrics, scoring the entire network.

\subsection{Jacobian Covariance}
    Jacobian Covariance is a training-free NAS metric for CNN networks proposed by \citet{mellor_neural_2021-1}. Given a minibatch of input data, the metric assesses the Jacobian of the network's loss function with respect to the minibatch inputs, $\mathbf{J} = \left(\frac{\partial\mathcal{L}}{\partial x_1}\cdots\frac{\partial\mathcal{L}}{\partial x_N}\right)$. Further details of the metric can be found in the original paper.
    
    \citet{celotti_improving_2020} expand on Jacobian Covariance with a series of variations on the metric, aiming to speed up computation and refine the metric's effectiveness. These include using cosine similarity instead of a covariance matrix to calculate similarity (Jacobian Cosine), 
    \begin{align*}
        S = 1 - \frac{1}{N^2-N}\sum_{i=1}^N\left|J_nJ_n^t - I\right|^{\frac{1}{20}},
    \end{align*}
    where $J_n$ is the normalized Jacobian and $I$ is the identity matrix, with a minibatch of $N$ inputs. In their Large Noise and More Noised scores, they add various noise levels to the input minibatch, hypothesizing that an architecture with high accuracy will be robust against noise.

\subsection{Synaptic Saliency}
    In the area of network pruning, \citet{tanaka_pruning_2020} proposed synaptic saliency, a score for approximating the change in loss when a specific parameter is removed. Synaptic saliency is based on the idea of preventing  layer collapse while pruning a network, which significantly decreases the network's accuracy. Synaptic saliency is expressed by
    \begin{align}
        S(\theta) = \frac{\partial\mathcal{L}}{\partial\theta} \odot \theta,
    \end{align}
    where $\mathcal{L}$ is the loss function, $\theta$ is the network's parameters, and $\odot$ is the Hadamard product. \citet{abdelfattah_zero-cost_2020} generalize synaptic saliency as a training-free metric for NAS by summing over all $N$ parameters in the network: $S = \sum_{i=1}^N S(\theta_i)$. \citet{abdelfattah_zero-cost_2020} found that synaptic saliency slightly outperforms Jacobian covariance on the NAS-Bench-201 CNN benchmark.

\subsection{Activation Distance}
    In a revised version of their paper, \citet{mellor_neural_2021} developed a more efficient metric that directly looks at the ReLU activations of a network. Given a minibatch of inputs fed into the network, the metric calculates the similarity of the activations within the initialized network between each input using their Hamming distance. \citeauthor{mellor_neural_2021} conclude that the more similar the activation map for a given set of inputs are to each other, the harder it is for the network to disentangle the representations of the inputs during training.

\subsection{Synaptic Diversity}
    \citet{zhou_training-free_2022} developed a metric specific for vision transformers (ViT)~\cite{dosovitskiy_image_2021}. Synaptic diversity is based upon previous research on rank collapse in transformers, where for a set of inputs the output of a multi-headed attention block converges to rank 1, significantly harming the performance of the transformer. \citeauthor{zhou_training-free_2022} use the Nuclear-norm of an attention heads's weight matrix $W_m$ as an approximation of its rank, creating the synaptic diversity score:
    \begin{align*}
        S_D = \sum_m\left|\left|\frac{\partial\mathcal{L}}{\partial W_m}\right|\right|_{nuc}\odot||W_m||_{nuc}.
    \end{align*}

\subsection{Hidden Covariance}
    We propose a new metric specific for RNNs, based on the hidden states between each layer of the RNN architecture. Previous NAS metrics focus on either the activation functions within an architecture, or all parameters of the architecture. The hidden state of an RNN layer encodes all of the information of the input, before being passed to the next layer or the final output. We hypothesize that if the hidden states of an architecture given a minibatch of inputs are similar to each other, the more difficult it would be to train the architecture, similar to \citet{mellor_neural_2021}.

    Given the hidden state $\mathbf{H}(\mathbf{X})$ of a specific layer of the RNN with a minibatch of $N$ inputs $\mathbf{X} = \{\mathbf{x}_n\}_{n=1}^N$, observe the covariance matrix to be 
    \begin{align*}
        \mathbf{C} = (\mathbf{H}-\mathbf{M}_\mathbf{H})(\mathbf{H}-\mathbf{M}_\mathbf{H})^T,
    \end{align*} where $\mathbf{M}_\mathbf{H}$ is the matrix with the entries $(\mathbf{M}_\mathbf{H})_{ij} = \frac{1}{N}\sum_{n=1}^N\mathbf{H}_{in}$. Then, calculate the Pearson product-moment correlation coefficients matrix
    \begin{align*}
        \mathbf{R}_{ij} = \frac{\mathbf{C}_{ij}}{\sqrt{\mathbf{C}_{ii}\mathbf{C}_{jj}}}.
    \end{align*}
    As with \citeauthor{mellor_neural_2021-1}'s Jacobian Covariance score (\citeyear{mellor_neural_2021-1}), the final metric is calculated with the Kullback–Leibler divergence of the kernel of $\mathbf{R}$, which has the $N$ eigenvalues $\lambda_1, \cdots, \lambda_N$:
    \begin{align*}
        S(\mathbf{H}) = -\sum_{n=1}^N\left(\log(\lambda_n + k) + \frac{1}{\lambda_n + k}\right),
    \end{align*}
    where $k = 10^{-5}$.

\subsection{Attention Confidence, Importance, and Softmax Confidence}
    For transformer-specific metrics, we look into current transformer pruning literature. \citet{voita_analyzing_2019} propose pruning the attention heads of a trained transformer encoder block by computing the ``confidence'' of a head using a sample minibatch of input tokens. Confident heads attend their output highly to a single token, and, hypothetically, are more important to the transformer's task. \citet{behnke_losing_2020} attempt to improve on attention confidence by looking at the probability distribution provided by an attention head's softmax layer. Alternatively, \citet{michel_are_2019} look at the sensitivity of an attention head to its weights being masked, by computing the product between the output of an attention head with the gradient of its weights. These three attention scores are summarized by:
    \begin{align*}
        \text{Confidence:}\ A_h(\mathbf{X}) &= \frac{1}{N}\sum_{n=1}^N\left|\max(\text{Att}_h(\mathbf{x}_n))\right| \\
        \genfrac{}{}{0pt}{}{\text{Softmax}}{\text{Confidence}}\text{:}\ A_h(\mathbf{X}) &= \frac{1}{N}\sum_{n=1}^N\left|\max(\sigma_h(\mathbf{x}_n))\right| \\
        \text{Importance:}\ A_h(\mathbf{X}) &= \left|\text{Att}_h(\mathbf{X})\frac{\partial\mathcal{L}(\mathbf{X})}{\partial\text{Att}_h(\mathbf{X})}\right|
    \end{align*}
    where $\mathbf{X} = \{\mathbf{x}_n\}_{n=1}^N$ is a minibatch of $N$ inputs, $\mathcal{L}$ is the loss function of the model, and $\text{Att}_h$ and $\sigma_h$ are an attention head and its softmax respectively. We expand these scores into an metric for the entire network by averaging over all $H$ attention heads: $\mathcal{A}(\mathbf{X}) = \sum_{h=1}^H\frac{1}{H}\text{Att}_h(\mathbf{X})$.

\section{Methods}

\subsection{NAS Benchmarks}
    Because of the large search space for neural architectures, it is challenging to have direct comparisons between various NAS algorithms. A series of NAS benchmarks~\cite{mehta_nas-bench-suite_2022} have been created, which evaluate a set of architectures within a given search space and store the trained metrics in a lookup table. These benchmarks include NAS-Bench-101~\cite{ying_nas-bench-101_2019}, NAS-Bench-201~\cite{dong_nas-bench-201_2020}, and NAS-Bench-301~\cite{siems_nas-bench-301_2021} with CNNs for image classification, NAS-Bench-ASR with convolutional LSTMs for automatic speech recognition~\cite{mehrotra_nas-bench-asr_2021}, and NAS-Bench-NLP with RNNs for language modeling tasks~\cite{klyuchnikov_nas-bench-nlp_2022}. Because the architectures in a NAS benchmark have already been trained, they allow for easier development of NAS algorithms without the large amounts of computational power required to train thousands of architectures. There are no existing NAS benchmarks for transformer or BERT-based architectures, due to the longer time and higher computing power required to train transformers.

    To evaluate training-free metrics on RNNs, we utilize the NAS-Bench-NLP benchmark~\cite{klyuchnikov_nas-bench-nlp_2022}, which consists of 14,322 RNN architectures trained for language modeling with the Penn Treebank dataset~\cite{marcus_building_1993}, each with precomputed loss values. The architecture search space is defined by the operations within an RNN cell, connected in the form of an acyclic digraph. The RNN architecture consists of three identical stacked cells with an input embedding and connected output layer. Further details on the architectures are provided in \citeauthor{klyuchnikov_nas-bench-nlp_2022}'s paper. In our experiments, the architectures which did not complete training within the benchmark or whose metrics could not be calculated were discarded, leaving 8,795 architectures that were evaluated on.

    \begin{table*}[h]
        \centering 
        \begin{tabular}{|l|l|}
        \hline
        \textbf{Architecture Element} & \textbf{Hyperparameters Values} \\
        \hline
        Hidden dimension & \{128, 256\} \\
        \hline
        Number of Encoder Layers & \{2, 4\} \\
        \hline
        \hline
        Type of attention operator & \{self-attention, linear transform, span-based dynamic convolution\} \\
        \hline
        Number of operation heads & \{2, 4\} \\
        \hline
        Feed-forward dimension & \{512, 1024\} \\
        \hline
        Number of feed-forward stacks & \{1, 3\} \\
        \hline
        Attention operation parameters & \\
        \quad if self-attention & \{scaled dot-product, multiplicative\} \\
        \quad if linear transform & \{discrete Fourier, discrete cosine\} \\
        \quad if dynamic convolution & \ convolution kernel size: \{5, 9\} \\
        \hline
        \end{tabular}
        \caption{The FlexiBERT search space, with hyperparameter values spanning those found in BERT-Tiny and BERT-Mini. Hidden dimension and number of encoder layers is fixed across the whole architecture; all other parameters are heterogeneous across encoder layers. The search space encompasses 10,621,440 architectures.}
         \label{table1}
    \end{table*}

\subsection{BERT Benchmark for NAS}
    Because no preexisting NAS benchmark exists for BERT-based architectures, we needed to pretrain and evaluate a large set of various BERT architectures in order to evaluate our proposed training-free NAS metrics. Certain choices were made in order to speed up pretraining while preserving relative model performance. These included: using the ELECTRA pretraining scheme~\cite{clark_electra_2020}, choosing a search space consisting of small BERT architectures, and shortening pretraining. 

\subsubsection{BERT Search Space}
    BERT (Bidirectional Encoder Representations from Transformers)~\cite{devlin_bert_2019} consists of a series of encoder layers with multi-headed self-attention, taken from the original transformer model proposed by \citet{vaswani_attention_2017}. Numerous variations on the original BERT model have been developed. For our architecture search space, we utilize the FlexiBERT search space~\cite{tuli_flexibert_2022}, which has improvements over other proposed BERT search spaces. Foremost is that the encoder layers in FlexiBERT are heterogeneous, each having their own set of architecture elements. FlexiBERT also incorporates alternatives to the multi-headed self-attention into its search space. The search space is described in Table \ref{table1}. 
    
    The architectures in the FlexiBERT search space are relatively small, as the hyperparameter values in the FlexiBERT search space spans those in BERT-Tiny and BERT-Mini~\cite{turc_well-read_2019}. However, \citet{kaplan_scaling_2020} show that many attributes of a transformer architecture, including number of parameters, scale linearly with the architecture's performance. Thus, a transformer architecture can be scaled up in order to achieve greater performance while preserving its overall structure. This methodology was utilized in EcoNAS algorithm~\cite{zhou_econas_2020}, which explores a reduced search space, before scaling up to produce the final model.
    
    To allow for simpler implementation of the FlexiBERT search space and the utilization of absolute positional encoding, we keep the hidden dimension constant across all encoder layers. In total, this search space encompasses 10,621,440 different transformer architectures.

\subsubsection{ELECTRA Pretraining}
    Instead of the traditional masked language modeling (MLM) task used to pretrain BERT-based models, we implemented the ELECTRA pretraining scheme~\cite{clark_electra_2020}, which uses a combination generator-discriminator model with a replaced token detection task. As the \mbox{ELECTRA} task is defined over all input tokens, instead of only the masked tokens as in MLM, it is significantly more compute efficient and results in better finetuning performance when compared to masked-language modeling. Notably, ELECTRA scales well with small amounts of compute, allowing for efficient pretraining of small BERT models.

\subsubsection{Architecture Training and Evaluation}
    We pretrain a random sample of 500 architectures from the \mbox{FlexiBERT} subspace using ELECTRA with the OpenWebText corpus, consisting of 38 GB of tokenized text data from 8,013,769 documents~\cite{gokaslan_openwebtext_2019}. OpenWebText is an open-sourced reproduction of OpenAI's WebText dataset~\cite{radford_language_2019}. We finetune and evaluate the architectures on the General Language Understanding Evaluation (GLUE) benchmark~\cite{wang_glue_2019}, without the WNLI task. The hyperparameters used for pretraining and finetuning are the same as those used for ELECTRA-Small. The sampled architectures were only pretrained for 100,000 steps for the best trade-off between pretraining time and GLUE score. Further details are discussed in the Appendix.

    \begin{figure*}[h!]
        \centering
        \includegraphics[width=1\textwidth]{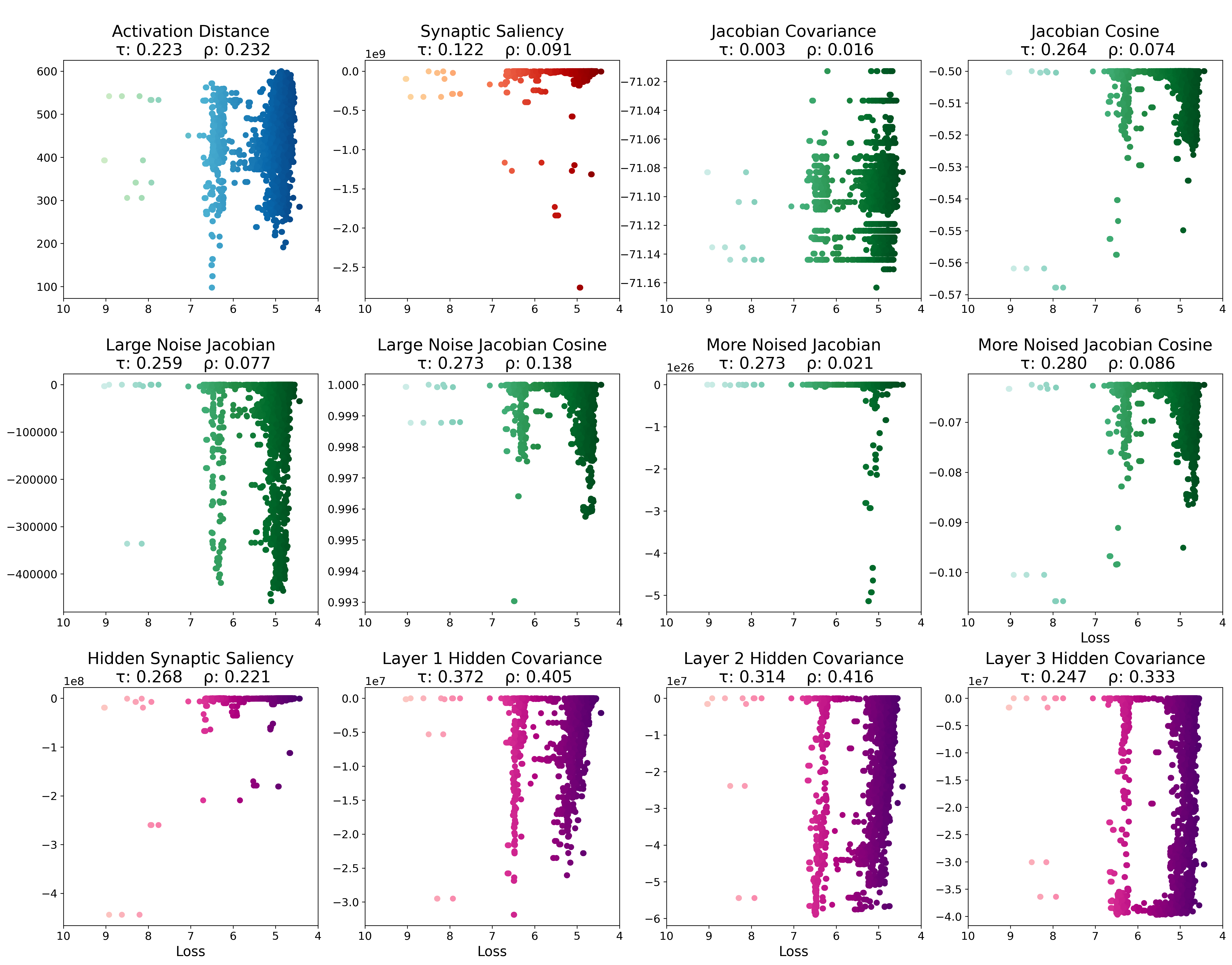}
        \caption{Plots of training-free metrics evaluated on 8,795 RNN architectures in NAS-Bench-NLP, against test loss of the architectures assessed on the Penn Treebank dataset when trained. Loss values are from NAS-Bench-NLP, and Kendall $\tau$ and Spearman $\rho$ also shown. Only our Hidden Covariance metric performed on the first and second layer of the RNN showed a substantial correlation between the metric and trained test loss. Some other metrics do have some minor positive correlations.}
        \label{fig1}
    \end{figure*}

\section{Experimental Results of Training-free Metrics}
    For the training-free NAS metrics presented, we empirically evaluate how well the metric performs in predicting the trained performance of an architecture. We use Kendall rank correlation coefficient (Kendall $\tau$) and Spearman rank correlation coefficient (Spearman $\rho$) to quantitatively measure the metrics' performance.

\subsection{Training-free Metrics for RNNs}
    We ran the training-free metrics on 8,795 architectures in NAS-Bench-NLP. A summary of our results are show in Figure \ref{fig1}. Most metrics preform poorly on predicting the loss of a trained RNN architecture, including all the existing training-free metrics designed for CNN architectures. No existing metric surpassed a Kendall $\tau$ value of $0.28$. Our proposed Hidden Covariance score preforms the best out of all metrics, achieving a Kendall $\tau$ value of $0.37$. Thus, the hidden states contain the most salient information for predicting the RNN's trained accuracy.

    \begin{figure*}[h!]
        \centering
        \includegraphics[width=\textwidth]{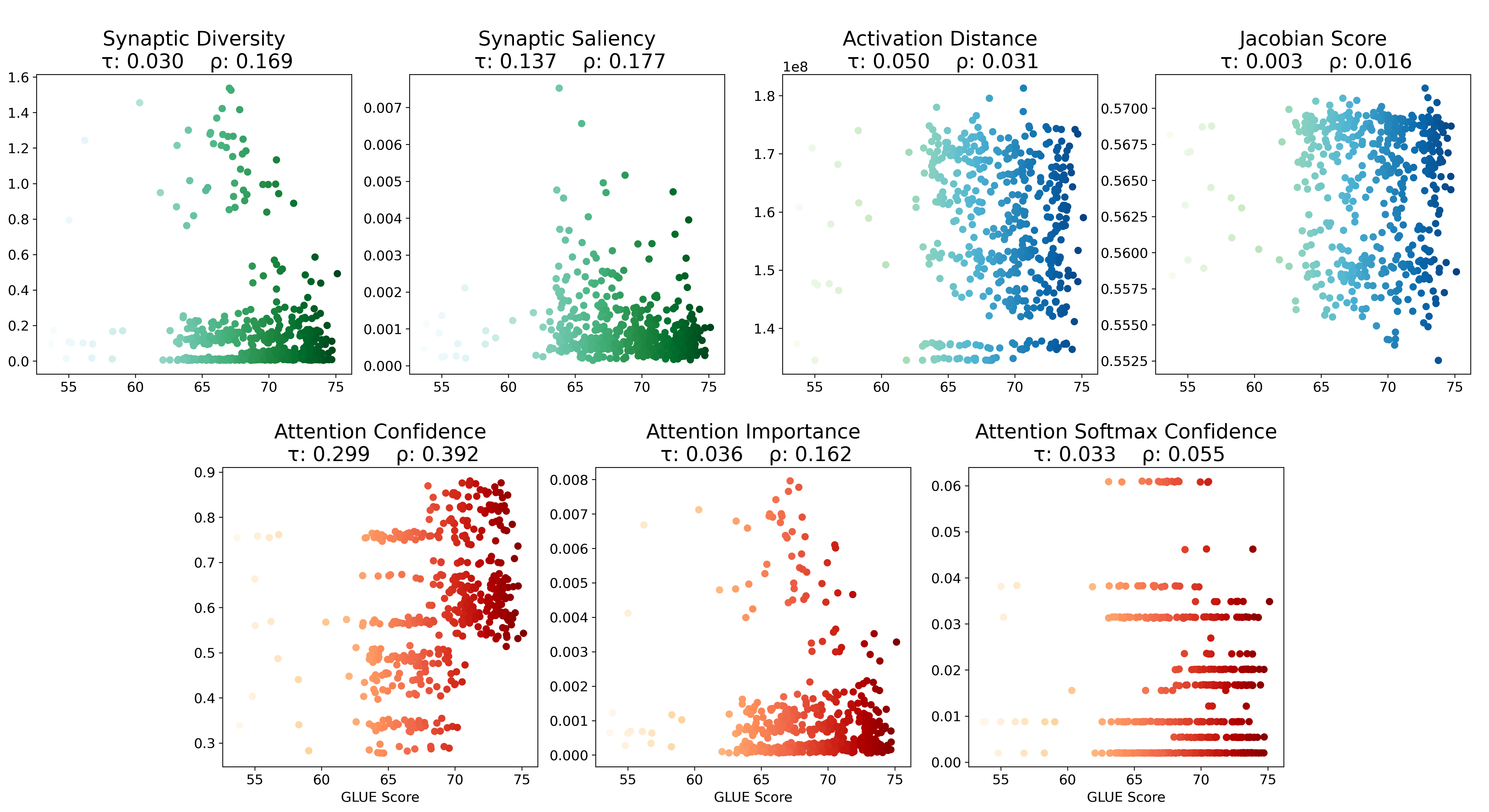}
        \caption{Plots of training-free metrics evaluated on 500 architectures randomly sampled from the FlexiBERT search space, against GLUE score of the pretrained and finetuned architecture. All metrics are normalized against number of features. Only our Attention Confidence metric displayed any positive correlation between the metric and final GLUE score.}
        \label{fig2}
    \end{figure*}

\subsection{Training-free Metrics for BERT Architectures}
    We investigated the series of training-free metrics on our own NAS BERT benchmark of 500 architectures sampled from the FlexiBERT search space. Results are shown in Figure \ref{fig2}. Compared to their performance on NAS-Bench-NLP, all the training-free metrics, including our proposed attention head pruning metrics, performed poorly. Only the Attention Confidence metric had a weak but significant positive correlation, with a Kendall $\tau$ of $0.27$.

    \begin{figure}[h!]
        \centering
        \includegraphics[width=0.9\columnwidth]{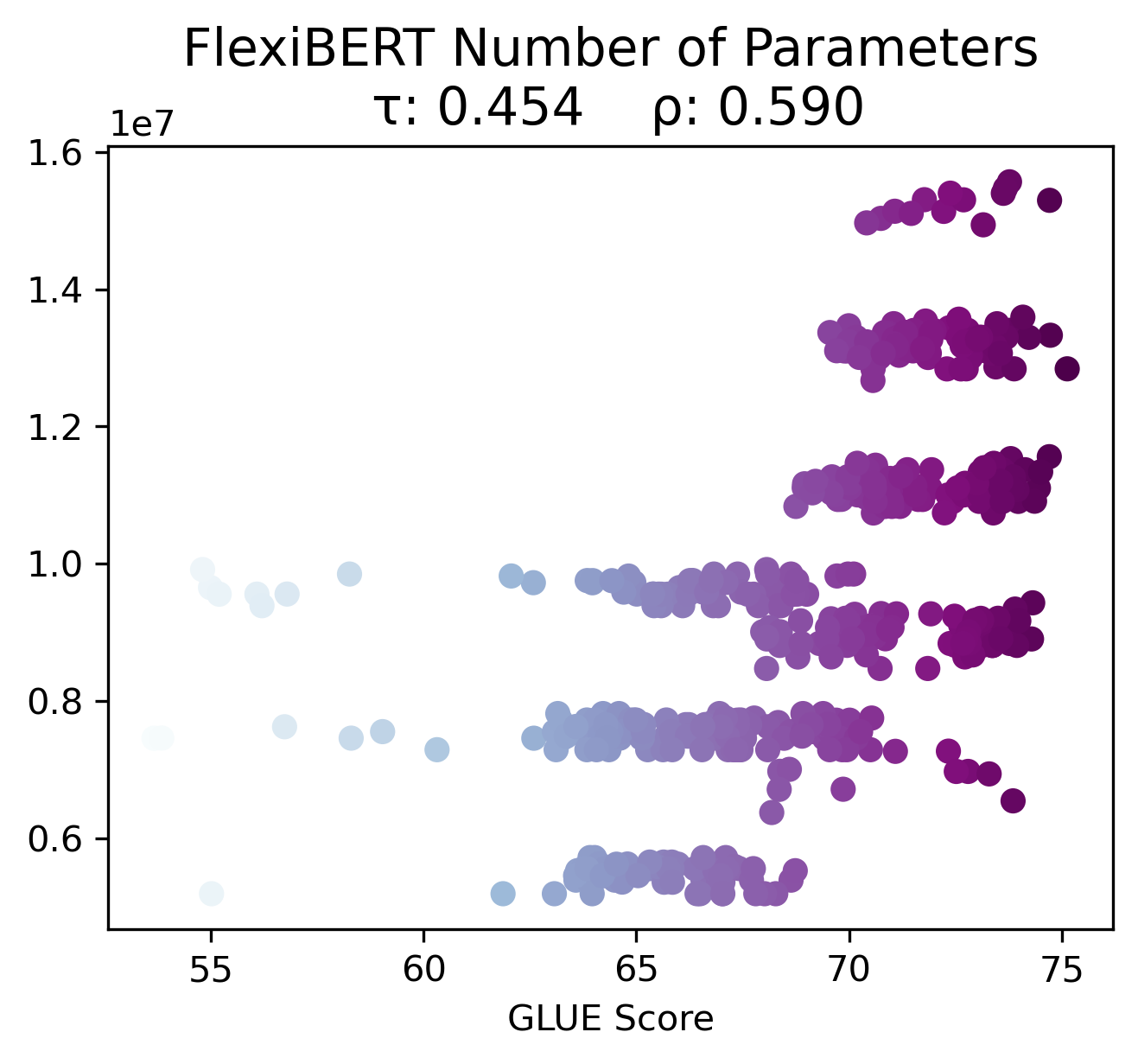}
        \caption{Correlation between number of parameters in a BERT-based architecture and its pretrained and finetuned GLUE score, for 500 architectures from the FlexiBERT search space. Number of parameters shows a strong correlation with architecture performance, substantially outperforms all training-free metrics evaluated.}
        \label{fig3}
    \end{figure}

    A notable reference point for training-free metrics is the number of trainable parameters in a transformer architecture. Previous research has shown a strong correlation between number of parameters and model performance across a wide range of transformer sizes and hyperparameters~\cite{kaplan_scaling_2020}. Our NAS BERT Benchmark displays this same correlation (Figure \ref{fig3}). In fact, the Kendall $\tau$ value for number of parameters is $0.44$, significantly surpassing all training-free metrics.

    \begin{figure}[h!]
        \centering
        \includegraphics[width=0.9\columnwidth]{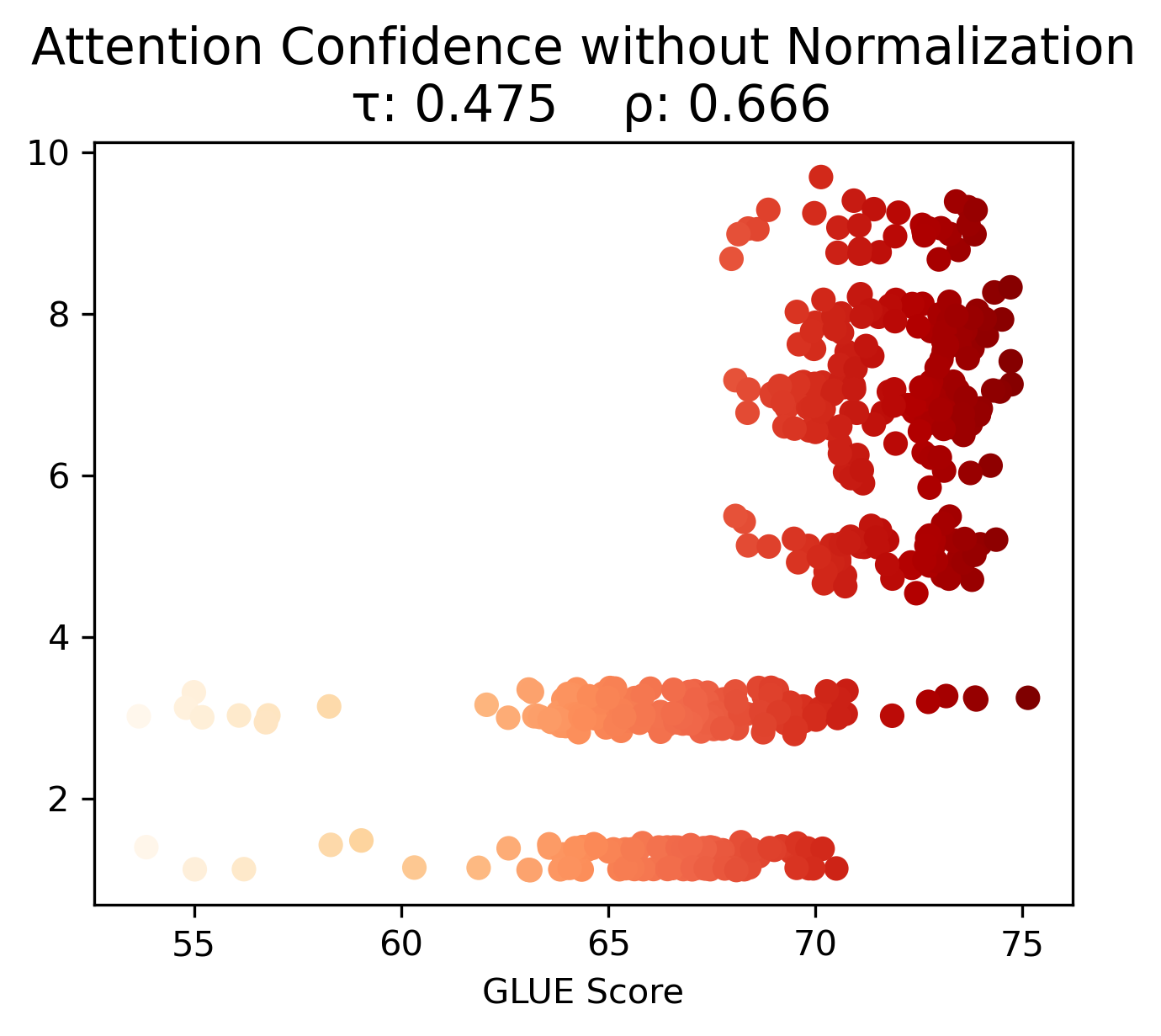}
        \caption{Attention Confidence metric evaluated on architectures from the FlexiBERT search space, without normalization for number of features. The metric's performance substantially improves when not normalized, and its plot and Kendall $\tau$ value mirrors that of number of parameters.}
        \label{fig5}
    \end{figure}

    Great care must be used when developing training-free metrics to ensure that the metric is normalized for number of parameters or other high-level features of the network. Many training-free metrics are computed on individual network features, which are then summed together to produce a final score for the network. In \citeauthor{zhou_training-free_2022}'s DSS-indicator score for vision transformers (a combination of synaptic saliency and synaptic diversity metrics), the score was not normalized for the number of features in the network (\citeyear{zhou_training-free_2022}). Instead, the DSS-indicator corresponds to the number of parameters in an architecture, as shown in their figures, thus yielding their high Kendall $\tau$ of $0.70$. We witnessed a similar pattern with our metrics. Attention Confidence had a Kendall $\tau$ of $0.49$ without normalization for number of features, but decreased to $0.30$ with normalization (Figure \ref{fig5}).

\section{Discussion}

\begin{figure}[h]
        \centering
        \includegraphics[width=0.9\columnwidth]{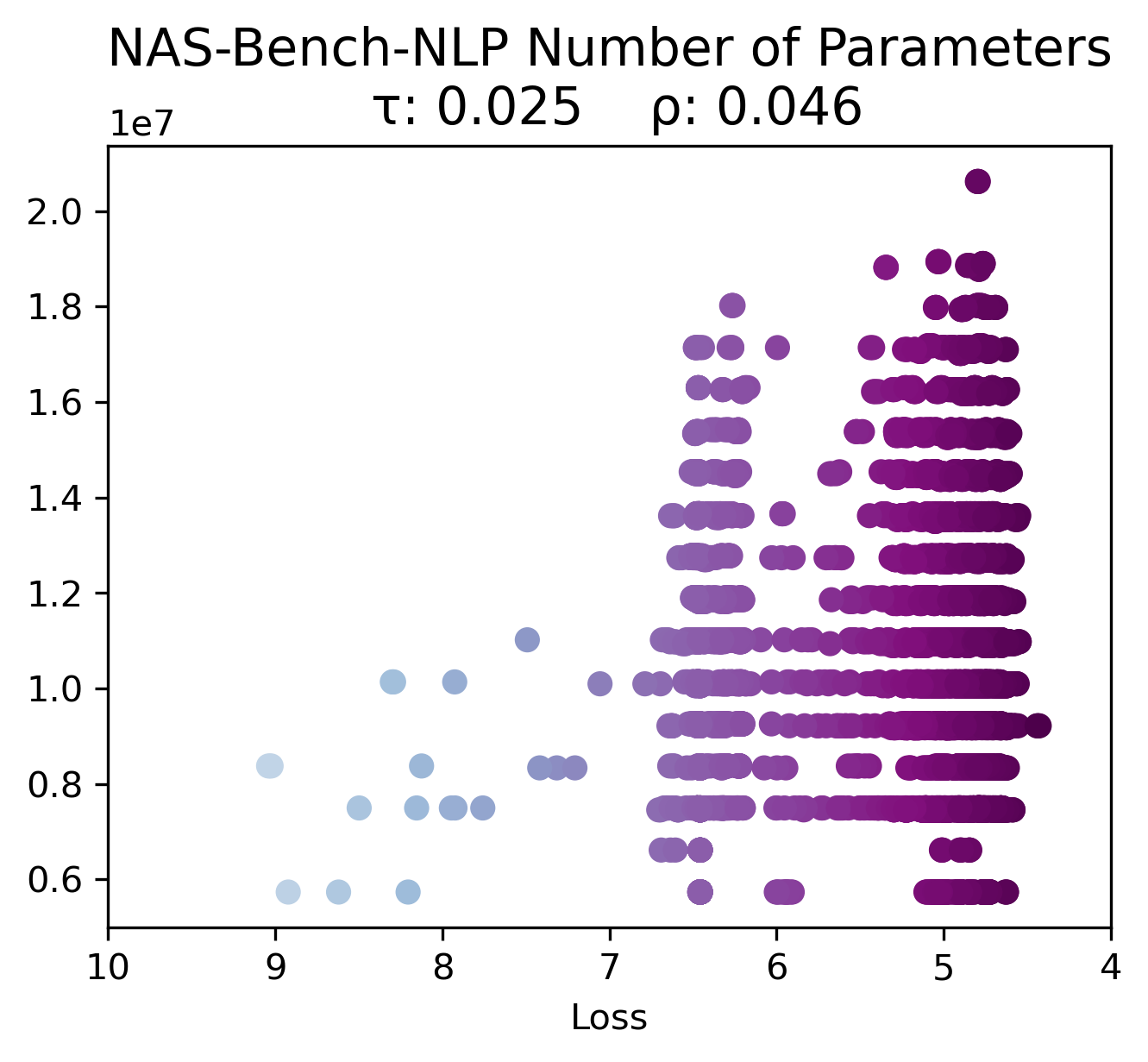}
        \caption{Plot of number of parameters against test loss for 
        8,795 RNN architectures in NAS-Bench-NLP. Unlike the architectures in the FlexiBERT search space, there is no correlation between number of parameters and architecture performance for the architectures in NAS-Bench-NLP.}
        \label{fig4}
    \end{figure}

    Neural architecture search for transformers is a fundamentally different task than neural architecture search for CNNs and RNNs. Almost all search spaces for transformers rely on the same fundamental paradigm of an attention module followed by a feed-forward module within each encoder/decoder layer, connected linearly~\cite{wang_hat_2020, yin_autotinybert_2021, zhao_memory-efficient_2021}. Conversely, most search spaces for CNNs and RNNs, including NAS-Bench-201 and NAS-Bench-NLP, use a cell-based method, typically with an acyclic digraph representing the connections between operations~\cite{dong_nas-bench-201_2020, jing_nasabn_2020, klyuchnikov_nas-bench-nlp_2022, tan_mnasnet_2019}, allowing for significantly more flexibility in cell variation. For CNN and RNN search spaces, the connections between operations within a cell have a greater impact on the architecture's performance than number of parameters. In NAS-Bench-NLP, there is no correlation between number of parameters and model performance (Figure \ref{fig4}); hence, previous studies did not need to normalize their training-free metrics for number of parameters or features. We hypothesize that for transformer search spaces, the number of parameters in an architecture dominates the model performance, explaining the poor performance for training-free NAS metrics.

    The dependence on number model size for transformer models reveals a significant problem regarding transformer architecture search: the inflexibility of current transformer search spaces. Unless transformer search spaces adopt the variability of connections provided by a cell-based methods, as used by CNN and RNN search spaces, simple heuristics such as number of parameters and features will be the primary training-free predictors of transformer model performance. To our knowledge, only three works have utilized cell-based methods for transformer search spaces, the original transformer architecture search paper, ``The Evolved Transformer'' by \citet{so_evolved_2019}, its successor ``Primer''~\cite{so_searching_2021}, and ``AutoBERT-ZERO''~\cite{gao_autobert-zero_2022}. Some research has been done with cell-based search spaces for Conformers~\cite{shi_efficient_2021} and Vision Transformers~\cite{guo_nat_2020}, but only on the convolution modules of the architectures. Ultimately, there is significant opportunity for growth regarding transformer architecture search, and with it training-free NAS metric for transformers.

\section{Conclusion}
    In this paper, we presented and evaluated a series of training-free NAS metrics for RNN and BERT-based transformer architectures, trained on language modeling tasks. We developed new training-free metrics targeted towards specific architectures, hidden covariance for RNNs, and three metrics based on attention head pruning for transformers. We first verified the training-free metrics on  NAS-Bench-NLP, and found our hidden covariance metric outperforms existing training-free metrics on RNNs. We then developed our own NAS benchmark for transformers within the FlexiBERT search space, utilizing the ELECTRA scheme to significantly speed up pretraining. Evaluating the training-free metrics on our benchmark, our proposed Attention Confidence metric performs the best. However, the current search space paradigm for transformers is not well-suited for training-free metrics, and the number of parameters within a model is the best predictor of transformer performance. Our research shows that training-free NAS metrics are not universally successful across all architectures, and better transformer search spaces should be developed for training-free metrics to succeed. We hope that our work is a foundation for further research into training-free metrics for RNNs and transformers, in order to develop better and more efficient NAS techniques.

\section{Limitations}
    In our paper, we presented existing and novel training-free NAS metrics for RNNs and transformers. Benchmarks are required to evaluate the effectiveness of these metrics on various architectures. While there exists a robust benchmark for RNN architectures (NAS-Bench-NLP), there is none for transformer models. Thus, we had to create our own NAS benchmark. For our work, we were limited by the computational resources available to us, so we were only able to pretrain and finetune 500 models for our NAS BERT benchmark. A larger sample size would give a more accurate evaluation of the training-free NAS metrics. Furthermore, we only investigated the FlexiBERT search space. While FlexiBERT has a diverse search space, having heterogeneous layers and alternative attention operators, the variation between possible architectures is limited and still dependent on the linear paradigm of BERT. Alternative transformer search spaces using cell-based methods, such as those presented in ``Primer''~\cite{so_searching_2021} and ``AutoBERT-ZERO''~\cite{gao_autobert-zero_2022}, do not have this limitation. We were ultimately unable to investigate the performance of training-free NAS metrics on this type of search space, as there are no available benchmarks for these search spaces, and their greater variability necessitates a copiously large sample size that is well outside our computational capabilities.

    Another limitation is that we only evaluated the effectiveness of the presented metrics on encoder-only transformer architectures, and not encoder-decoder or decoder-only architectures. Furthermore, while the training-free NAS metrics are data-agnostic, the benchmarks they were evaluated on were only trained and evaluated on English datasets and tasks.

\section{Ethics Statement}
    The work presented in our paper is dependent on existing open source datasets and benchmarks, including OpenWebText~\cite{gokaslan_openwebtext_2019}, NAS-Bench-NLP~\cite{klyuchnikov_nas-bench-nlp_2022}, and GLUE~\cite{wang_glue_2019}. Therefore, our work inherently contains the ethical issues and limitations present in them. However, the ethics of these datasets and benchmark are largely unknown (despite OpenWebText and GLUE being widely used), as they were released without model or dataset cards and their authors do not discuss the societal impacts of their work.

    In our work, we adhere to best practices for reproducibility and descriptive statistics by sufficiently documenting our experimental setup and parameters, sharing our code and benchmark, and conducting ablation studies. One concern is the environmental and energy impact of creating our NAS BERT benchmark through the computationally intensive task of training of 500 unique transformer architectures. We decreased the environmental impact of our benchmark by reducing the size of the architectures, utilizing the more computationally efficient ELECTRA scheme pretraining, and limiting pretraining to 100,000 steps. We hope that the environmental impact is mitigated by openly sharing the benchmark, and the potential for training-free NAS metrics to drastically speed up NAS algorithms. Because metrics and NAS benchmark presented in our work are largely for theoretical purposes and only aid the creation of new architectures through NAS algorithms, the risk for harmful effects and uses resulting directly from our work is minimal.

    The NAS-Bench-NLP~\cite{klyuchnikov_nas-bench-nlp_2022}, ELECTRA~\cite{clark_electra_2020}, and the HuggingFace implementation of ELECTRA are released under the Apache License 2.0, which permits for commercial and non-commercial use, distribution, and modification. While the contents of the OpenWebText corpus was scraped from public websites without consent, the packaging of the corpus is released into the public domain under the Creative Commons CC0 license. The creators of OpenWebText allow individuals to submit take down requests of their own copyrighted works in the corpus. The Penn Treebank dataset~\cite{marcus_building_1993} is released under the Linguistic Data Consortium User Agreement for Non-Members, which permits use of the dataset for non-commercial research only, without distribution. In our work and the distribution of our code and dataset, we abide by the intended use of the code and datasets that we utilized, consistent with the terms of their licenses. We distribute our code under the Apache License 2.0 and our dataset under the Creative Commons Attribution 4.0 International Public License. 

\clearpage

\bibliography{bibliography}
\bibliographystyle{acl_natbib}

\clearpage
\appendix

\section{NAS BERT Benchmark Training Details}
    In the development of our NAS BERT benchmark, we did not aim to highly optimize the performance of the architectures on GLUE tasks. The goal of our benchmark was to compare transformer architectures solely with each other using training-free metrics, not to achieve state-of-the-art results surpassing other architectures. We want to have a large enough sample size of transformer architectures, even with our constrained compute capability. Thus, we chose to only use one pretraining dataset (OpenWebText~\cite{gokaslan_openwebtext_2019}), no hyperparameter optimization (Section \ref{hyperparameters}), only a single finetuning run on the GLUE benchmark for each architecture, and a reduced number of pretraining steps (Section \ref{train_steps}). Even with our suboptimal training choices, the architectures in our benchmark achieve comparable GLUE scores to other BERT-based models of the same size \cite{tuli_flexibert_2022, turc_well-read_2019}.

    We used the GLUE benchmark as it is widely used to evaluated BERT-based and other language modeling architectures~\cite{wang_glue_2019} (see GLUE leaderboard). We did not evaluated on the WNLI task, as the creators of the GLUE benchmark found that no model exceeds an accuracy of $65.1\%$ due to improper labeling of the train/dev/test sets. The scores for each GLUE task are Spearman's rank correlation coefficient for STS, Matthews's correlation coefficient for CoLA, and accuracy for all other tasks. These scores were averaged together into the final GLUE score. All GLUE results are from the dev set.

    All transformer architectures were trained on TPUv2s with 8 cores and 64 GB of memory, using Google Collabortory. The entire process of pretraining and finetuning our benchmark took approximately 25 TPU days. Evaluation of training-free metrics occurred on 2.8 GHz Intel Cascade Lake processors with either 16 or 32 cores and 32 GB of memory.

\subsection{Hyperparameters} \label{hyperparameters}
    \begin{table}[h]
        \centering 
        \begin{tabular}{l l}
        \hline
        \textbf{Hyperparameter} &  \\
        \hline
        Generator Size Multiplier & $1\backslash4$ \\
        Mask Percentage & $15\%$ \\
        Training Steps & 100,000 \\
        Learning Rate Decay & Linear \\
        Warmup Steps & 10,000 \\
        Learning Rate & 5e-4 \\
        Adam $\epsilon$ & 1e-6 \\
        Adam $\beta_1$ & 0.9 \\
        Adam $\beta_2$ & 0.999 \\
        Dropout & 0.1 \\
        Weight Decay & 0.01 \\
        Train Batch Size & 128 \\
        Evaluation Batch Size & 128 \\
        Vocabulary Size & 30522 \\
        \hline
        \end{tabular}
        \caption{Pretraining hyperparameters used to pretrain all architectures in our NAS BERT benchmark. Same parameters as used to pretrain ELECTRA-Small, except for number of training steps.}
        \label{table2}
    \end{table}

    For pretraining and finetuning the architectures in our NAS BERT benchmark, we used the same hyperparameters as use to train ELECTRA-Small, except for number of training steps (further discussion in main paper and Appendix Section \ref{train_steps}). These hyperparameters are listed in Table \ref{table2} and Table \ref{table3}. 

    \begin{table}[h!]
        \centering 
        \begin{tabular}{l l}
        \hline
        \textbf{Hyperparameter} &  \\
        \hline
        Learning Rate & 3e-4 \\
        Adam $\epsilon$ & 1e-6 \\
        Adam $\beta_1$ & 0.9 \\
        Adam $\beta_2$ & 0.999 \\
        Learning Rate Decay & Linear \\
        Layerwise LR decay & 0.8 \\
        Warmup Fraction & 0.1 \\
        Attention Dropout & 0.1 \\
        Dropout & 0.1 \\
        Weight Decay & 0.01 \\
        Batch Size & 32 \\
        Vocabulary Size & 30522 \\
        Train Epochs & 10 for RTE and STS \\
        & 3 for all other tasks \\
        \hline
        \end{tabular}
        \caption{Finetuning hyperparameters used to finetune all architectures in our NAS BERT benchmark on all tasks in the GLUE benchmark. Same parameters as used to finetune ELECTRA-Small.}
        \label{table3}
    \end{table}

    \begin{figure}[h!]
        \centering
        \includegraphics[width=0.7\columnwidth]{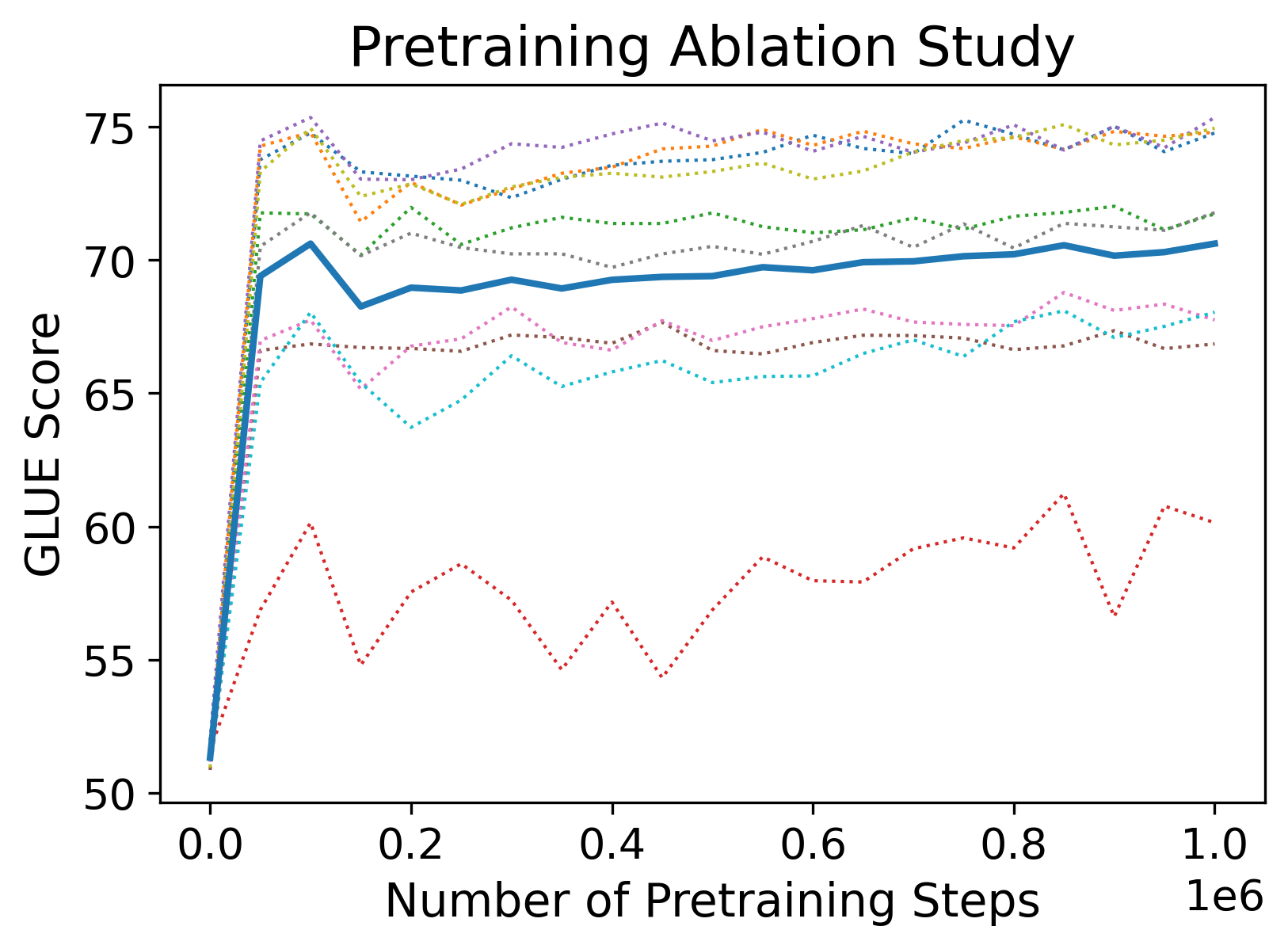}
        \caption{Pretraining ablation study of 10 architectures randomly sampled from the FlexiBERT search space investigating number of steps, using the hyperparameters in \ref{table3}. Dotted lines represent GLUE score of each individual architecture, and the blue line is average score of all architectures.}
        \label{fig6}
    \end{figure}

\subsection{Number of Training Steps} \label{train_steps}
    As discussed in Section 4.2.3 of the main paper, we chose to reduce the number of steps used for pretraining the architectures to be $100,000$, as opposed to the $1,000,000$ used to pretrain ELECTRA-Small. This choice was based on an ablation study of 10 architectures sampled from the benchmark (Figure \ref{fig6}). $100,000$ pretraining steps was determined to be the best trade-off between model performance on the GLUE benchmark and training time.

\section{Ablation Studies}
    Our evaluation of training-free metrics on both NAS-Bench-NLP and our NAS BERT benchmark requires random initialization of architectures, and many metrics require a mini-batch of input data, which we randomly sampled from respective datasets. To investigate the impact of initialization weights and input data, we conduct a series ablation studies for the training-free metrics on both benchmarks. 

\begin{figure*}[h!]
        \centering
        \includegraphics[width=1\textwidth]{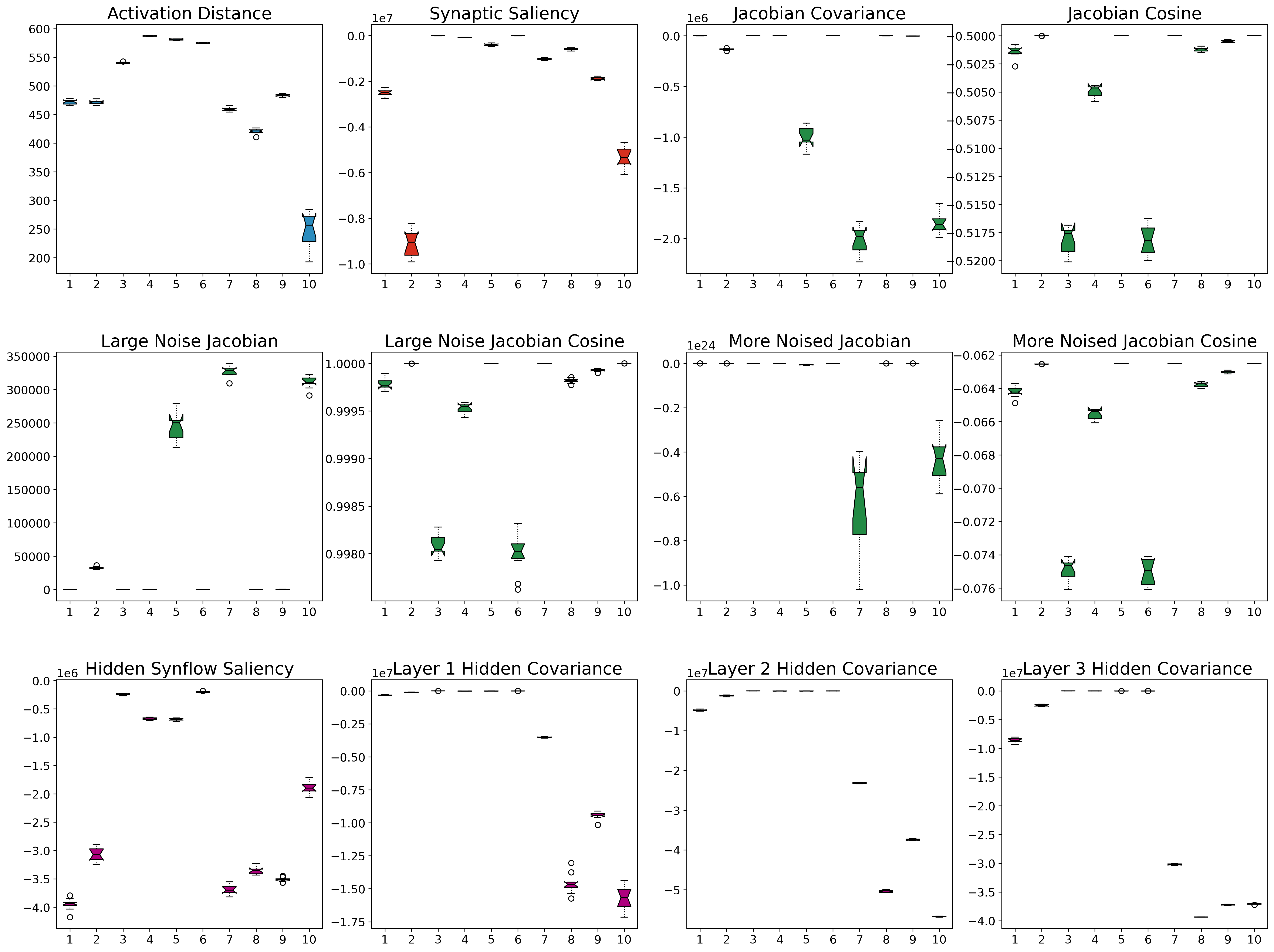}
        \caption{Ablation study showing the effect of different initialization weights on training-free metrics, evaluated using RNN architectures from NAS-Bench-NLP. 10 architectures were sampled from the benchmark, one in each decile range of test loss (eg. $0$-$10\%, 10$-$20\%, \dots, 90$-$100\%$). 10 different random seeds were used for the initialization weights.}
        \label{fig7}
    \end{figure*}

\begin{figure*}[h!]
        \centering
        \includegraphics[width=1\textwidth]{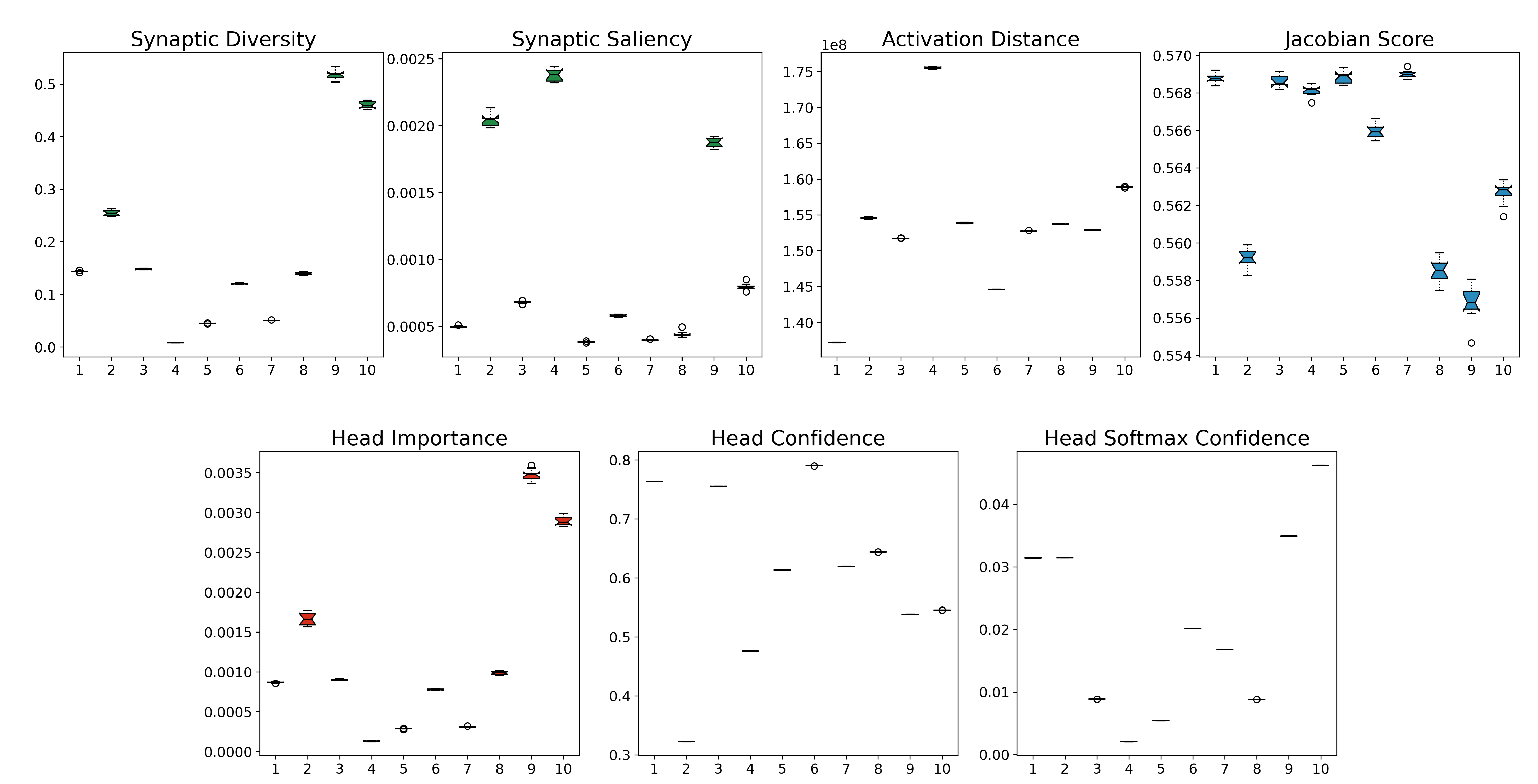}
        \caption{Ablation study showing the effect of different initialization weights on training-free metrics, evaluated using transformer architectures from our NAS BERT benchmark. 10 architectures were sampled from the benchmark, one in each decile range of GLUE score (eg. $0$-$10\%, 10$-$20\%, \dots, 90$-$100\%$). 10 different random seeds were used for the initialization weights.}
        \label{fig8}
    \end{figure*}

    Figures \ref{fig7} and \ref{fig8} show how the various training-free metrics evaluated on 10 architectures from NAS-Bench and our NAS BERT benchmark each differ with 10 different initialization weights. Overall, initialization weight has minimal impact on the evaluations of training-free metrics, and the metrics' scores are well distinguished between different architectures. While some metrics when evaluated on NAS-Bench-NLP architectures have larger variations, such as the More Noised Jacobian metric, the high performing metrics like Hidden Covariance can isolate better performing architectures. All metrics when evaluated on architectures from our NAS BERT benchmark have minimal variation between different initialization weights.

\begin{figure*}[h!]
        \centering
        \includegraphics[width=1\textwidth]{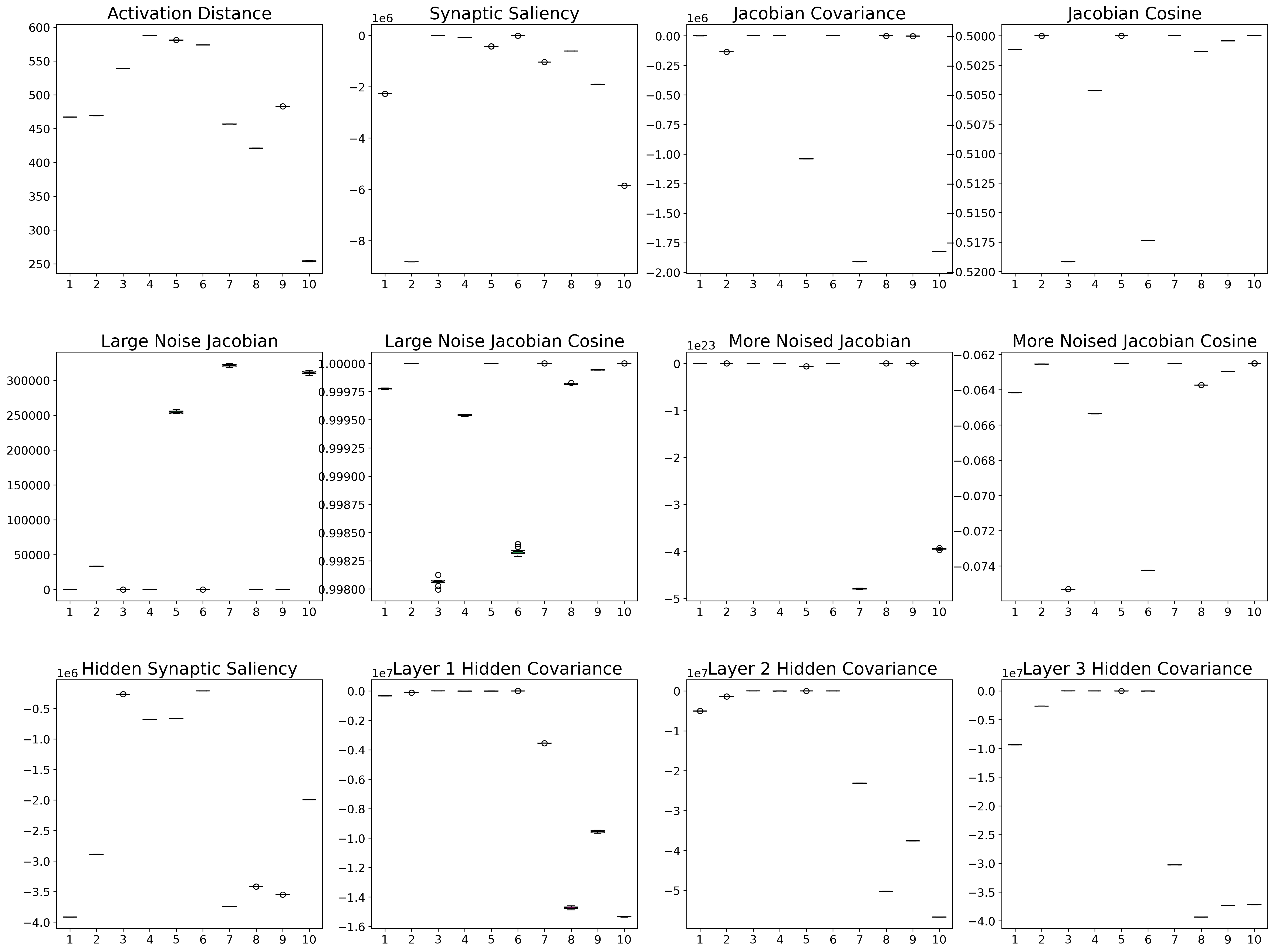}
        \caption{Ablation study showing the effect of different minibatch inputs on training-free metrics, evaluated using RNN architectures from NAS-Bench-NLP. 10 architectures were sampled from the benchmark, one in each decile range of test loss (eg. $0$-$10\%, 10$-$20\%, \dots, 90$-$100\%$). The same 10 minibatches of size 128, randomly selected from the Penn Treebank dataset, were used for each architecture and metric.}
        \label{fig9}
    \end{figure*}

\begin{figure*}[h!]
        \centering
        \includegraphics[width=1\textwidth]{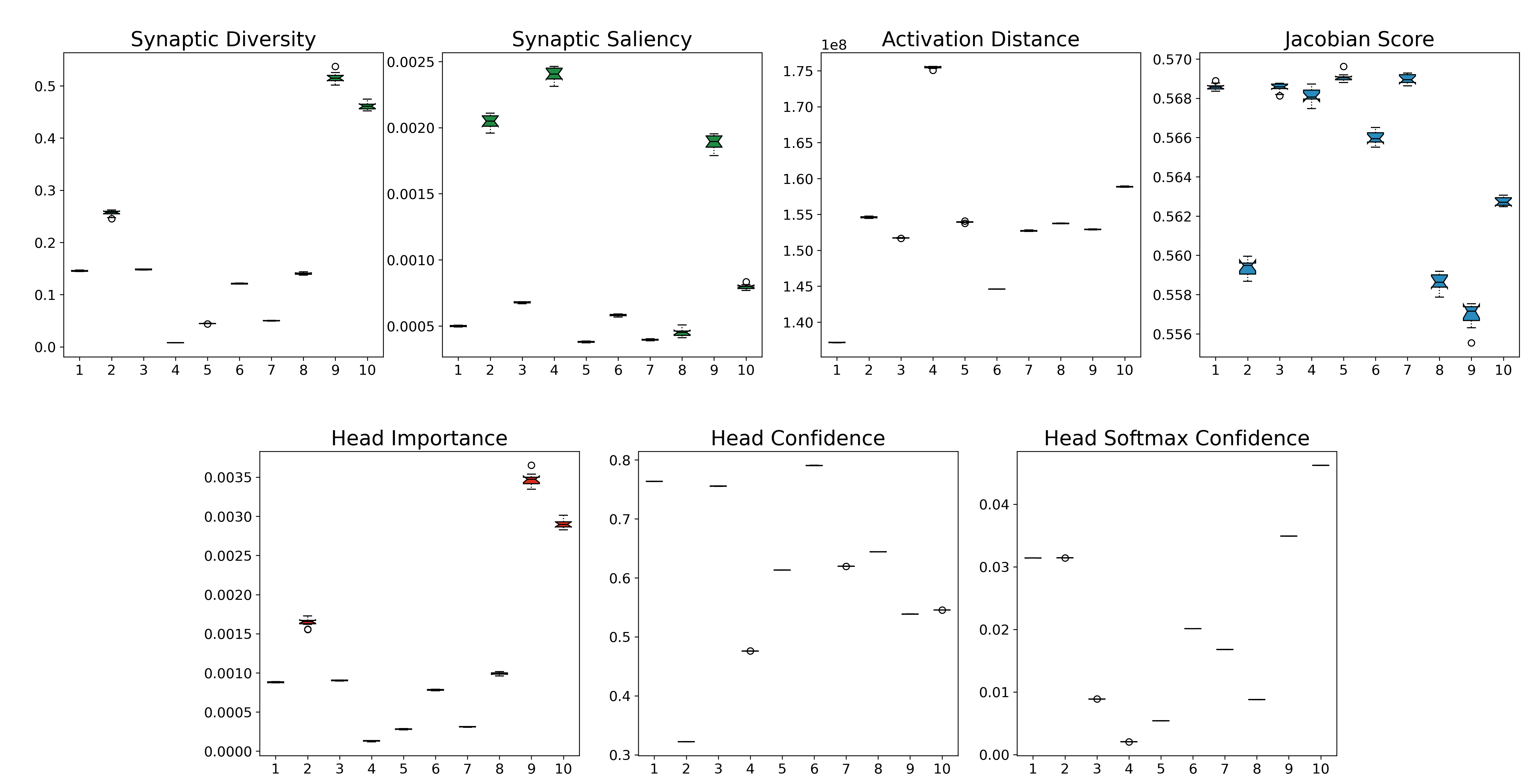}
        \caption{Ablation study showing the effect of different minibatch inputs on training-free metrics, evaluated using transformer architectures from our NAS BERT benchmark. 10 architectures were sampled from the benchmark, one in each decile range of test loss (eg. $0$-$10\%, 10$-$20\%, \dots, 90$-$100\%$). The same 10 minibatches of size 128, randomly selected from the OpenWebText dataset, were used for each architecture and metric.}
        \label{fig10}
    \end{figure*}

    Likewise, Figures \ref{fig9} and \ref{fig10} show the impact of 10 different input minibatches on training-free metrics. There is little variation in the metrics' evaluations for all metrics on both RNNs and BERT-based architectures.

    These ablation studies demonstrate that training-free metrics, when evaluated on RNN and transformer architectures, capture intrinsic properties contained within the architecture, rather than transient information in the specific input data or initialization.

\section{Non-Normalized Metrics on NAS BERT Benchmark}
\begin{figure*}[h!]
        \centering
        \includegraphics[width=1\textwidth]{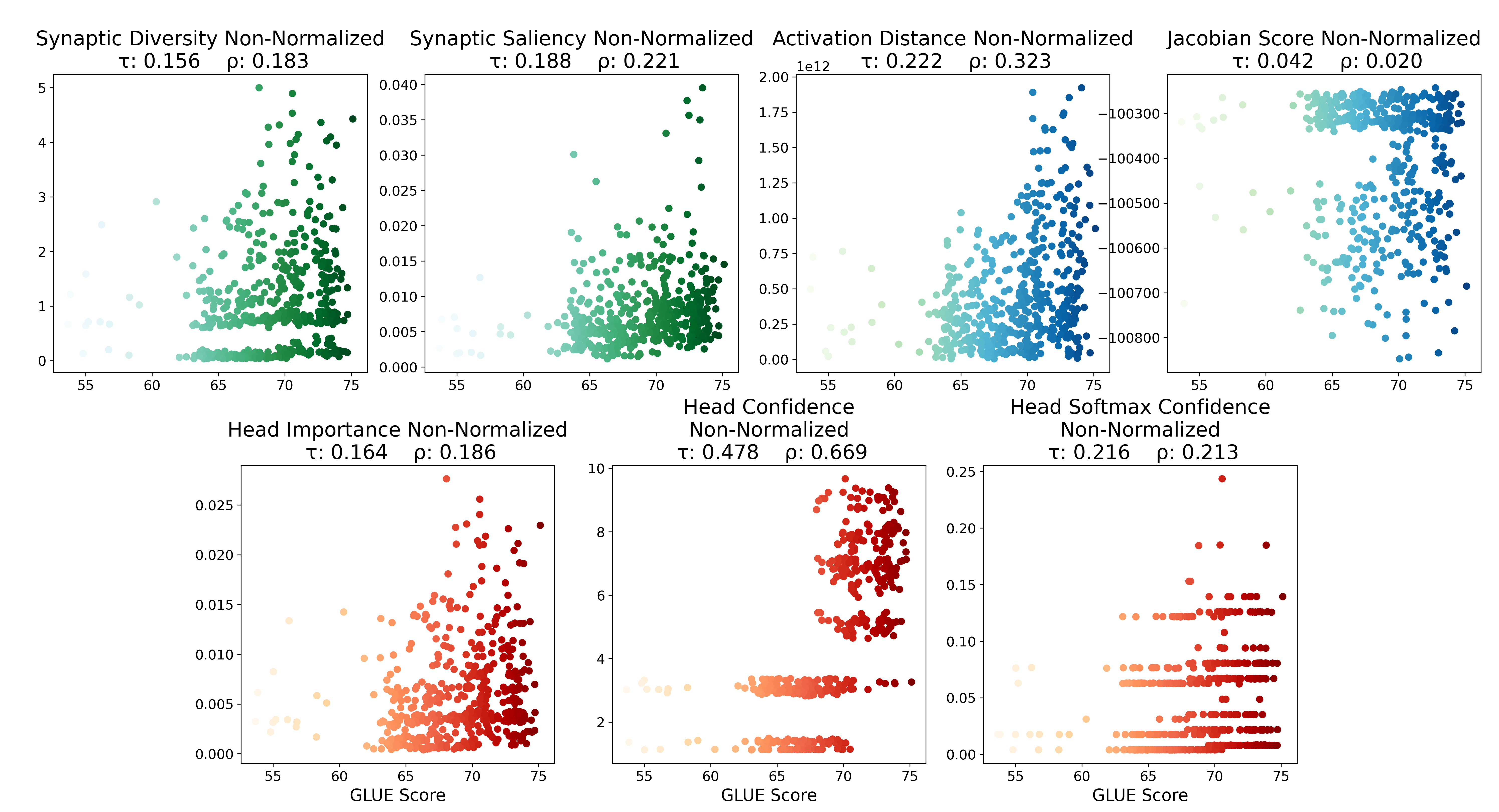}
        \caption{Plots of non-normalized training-free metrics evaluated on 500 architectures randomly sampled from the FlexiBERT search space, against GLUE score of the pretrained and finetuned architecture.}
        \label{fig11}
    \end{figure*}

    Continuing the discussion from Section 5.2 in the main paper, Figure \ref{fig11} shows the non-normalized training-free metrics when evaluated on our NAS BERT Benchmark. All metrics when not normalized for number of features increase in performance, with most showing some positive correlation. Head Confidence remains the best performing metric.

\end{document}